\documentclass{article}
\pdfoutput=1
\usepackage{microtype}
\usepackage{graphicx}
\usepackage{subfig} 
\usepackage{caption}
\usepackage{booktabs} 

\usepackage{hyperref}

\usepackage{color}
\usepackage{bm}
\usepackage{hyperref}
\usepackage{amsmath,amsfonts,amsthm}
\usepackage{blindtext}
\usepackage{listings}
\usepackage{algorithm, algorithmic}
\usepackage{booktabs}
\usepackage{multirow}
\usepackage{multicol}
\usepackage{caption}

\usepackage{ifthen}

\newif\ifdebug
\debugfalse

\newcommand{\vect}[1]{\boldsymbol{\mathbf{#1}}}

\newcommand{\dv}{\vect d}

\newcommand{\sv}{\vect s}

\newcommand{\Av}{\vect A}
\newcommand{\Bv}{\vect B}
\newcommand{\Cv}{\vect C}

\newcommand{\Fv}{\vect F}

\newcommand{\Sv}{\vect S}

\newcommand{\Wv}{\vect W}
\newcommand{\Xv}{\vect X}
\newcommand{\Yv}{\vect Y}

\newcommand{\Rb}{\mathbb R}

\makeatletter
\newtheorem*{rep@theorem}{\rep@title}
\newcommand{\newreptheorem}[2]{%
	\newenvironment{rep#1}[1]{%
		\def\rep@title{#2 \ref{##1}}%
		\begin{rep@theorem}}%
		{\end{rep@theorem}}}
\makeatother

\usepackage[accepted]{icml2024}
\usepackage{amsmath}
\usepackage{amssymb}
\usepackage{mathtools}
\usepackage{amsthm}

\usepackage[capitalize,noabbrev]{cleveref}

\theoremstyle{plain}

\theoremstyle{definition}

\theoremstyle{remark}

\usepackage[textsize=tiny]{todonotes}

\icmltitlerunning{Jetfire: Efficient and Accurate Transformer Pretraining with INT8 Data Flow and Per-Block Quantization}

\begin{document}

\twocolumn[
\icmltitle{Jetfire: Efficient and Accurate Transformer Pretraining with \\ INT8 Data Flow and Per-Block Quantization}



\icmlsetsymbol{intern}{*}

\begin{icmlauthorlist}
\icmlauthor{Haocheng Xi}{cs,iiis,intern}
\icmlauthor{Yuxiang Chen}{cs}
\icmlauthor{Kang Zhao}{cs}
\icmlauthor{KAI JUN TEH}{cs}
\icmlauthor{Jianfei Chen}{cs}
\icmlauthor{Jun Zhu}{cs}
\end{icmlauthorlist}

\icmlaffiliation{cs}{Dept. of Comp. Sci. and Tech., Institute for AI, BNRist Center, THBI Lab, Tsinghua-Bosch Joint ML Center, Tsinghua University}
\icmlaffiliation{iiis}{Institute for Interdisciplinary Information Sciences, Tsinghua University}

\icmlcorrespondingauthor{Jianfei Chen}{jianfeic@tsinghua.edu.cn}

\icmlkeywords{Machine Learning, ICML}

\vskip 0.3in]



\printAffiliationsAndNotice{\icmlEqualContribution} 

\begin{abstract}\label{Sec: abstract}
Pretraining transformers are generally time-consuming. Fully quantized training (FQT) is a promising approach to speed up pretraining. However, most FQT methods adopt a quantize-compute-dequantize procedure, which often leads to suboptimal speedup and significant performance degradation when used in transformers due to the high memory access overheads and low-precision computations. 

In this work, we propose Jetfire, an efficient and accurate INT8 training method specific to transformers. Our method features an INT8 data flow to optimize memory access and a per-block quantization method to maintain the accuracy of pretrained transformers. Extensive experiments demonstrate that our INT8 FQT method achieves comparable accuracy to the FP16 training baseline and outperforms the existing INT8 training works for transformers. Moreover,  for a standard transformer block, our method offers an end-to-end training speedup of 1.42x and a 1.49x memory reduction compared to the FP16 baseline. Our code is open sourced at \href{https://github.com/thu-ml/Jetfire-INT8Training}{https://github.com/thu-ml/Jetfire-INT8Training}.

\end{abstract}
\section{Introduction}\label{Sec: Intro}
Recently, large-scale pre-trained transformer-based models such as  GPT-4~\cite{GPT4}, LLAMA~\cite{LLAMA}, and PaLM~\cite{PALM2} have attained significant breakthroughs in multiple fields, including natural language processing and computer vision. However, pre-training transformers from scratch are extremely resource-intensive since they require numerous computations and high-bandwidth memory for updating weights and accessing huge amounts of training tokens, respectively.

\begin{figure}[t]
    \centering
    \includegraphics[width=0.9\linewidth]{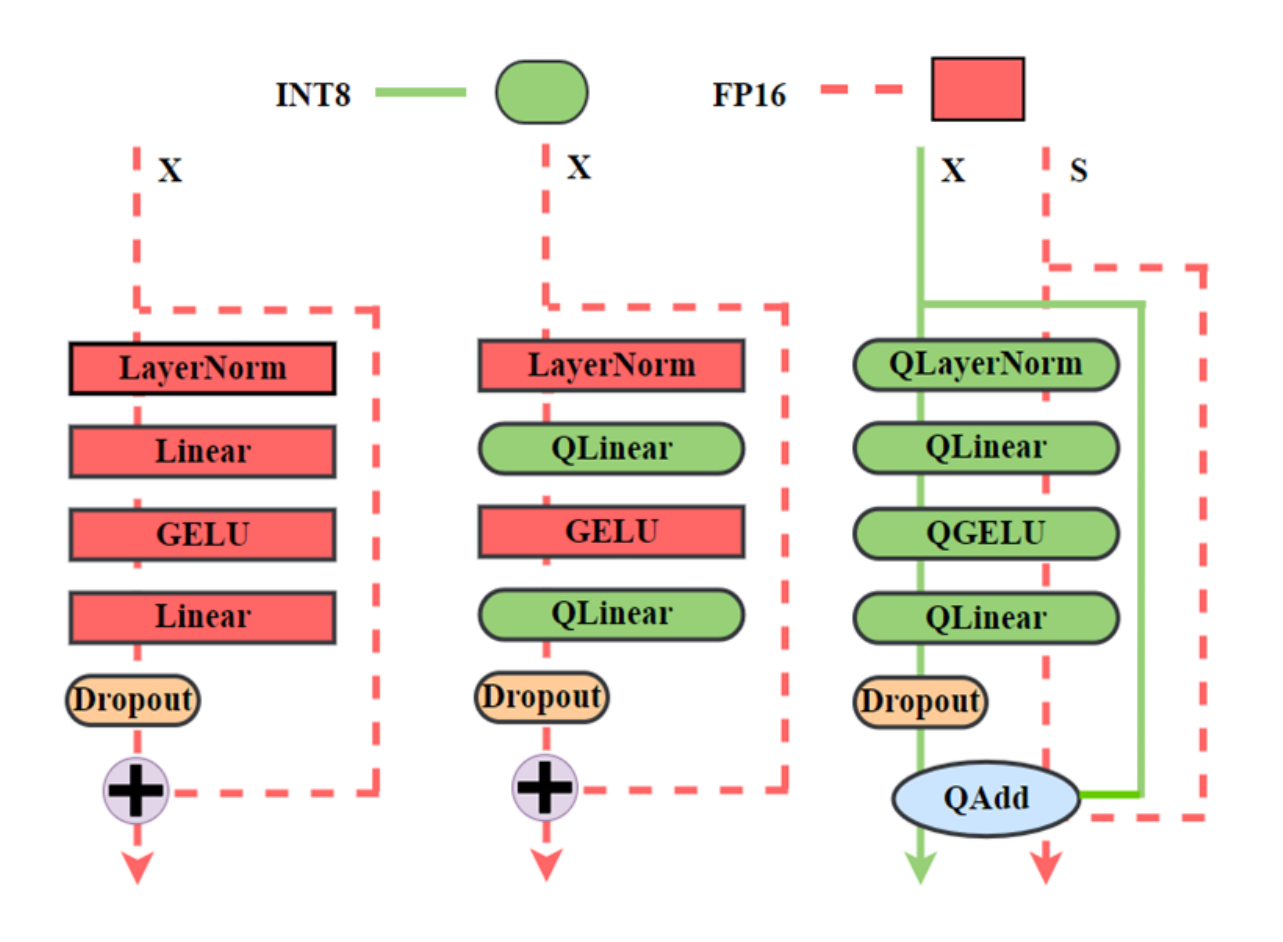}
    \caption{Visualization of INT8 data flow. (a) Floating point training with FP data flow. (b) Existing works on quantized training with FP data flow. (c) Ours INT8 training forward process, with INT8 data flow.
    $\Xv$ refers to the activation, and $\Sv$ refers to the corresponding quantization scale factors.}
    \label{Fig: INT8 training pipeline}
    \vspace{-0.4cm}
\end{figure}

To accelerate the pre-training of transformers, fully quantized training (FQT) has emerged as a promising technique to speed up both the forward and backward passes. FQT integrates quantizers and dequantizers into the original full-precision computational graph. In this way, the expensive floating-point operations during training are replaced with cheaper low-precision alternatives, and activations saved for the backward pass are stored with fewer bits.  Thus, both computations and memory bandwidths are largely reduced.  Typical FQT works include \cite{ 2018INT8, 2018FP8, AMP, chen2020statquant, INT8CLIP, INT4Train, ULTRALOW, LUQ, HFP8}.

However, the existing FQT studies still have three limitations: 
1) Existing FQT methods are not accurate enough for Transformer models. Previous FQT methods were mainly designed for CNNs \cite{INT8CNN, INT8Distributed}, and directly applying these methods to transformer models will result in significant accuracy degradation. Those few papers that focus on transformers often encounter significant quantization errors when computing weight gradients. Therefore, they leave this part in floating-point precision, which limits its overall speedup.
2) Most FQT methods only focus on the reduction of computations, regardless of data access overheads \cite{INT8CLIP}. Nevertheless, for a transformer block, only the linear layers are compute-bounded; other layers, such as LayerNorm and activation functions, are normally memory-bounded. Failing to optimize the memory access leads to suboptimal training acceleration. 
3) Some FQT techniques require specialized hardware and are not applicable to general computing platforms.  For instance, FP8~\cite{FP8LM, FP8graphcore} training is only supported on GPUs with Hopper architecture ~\cite{nvidiaH100}. Not to mention that those hybrid-format quantized training methods rely on application-specific integrated circuits to deliver the desired performance. 

To address these limitations, in this work, we propose Jetfire, 
an INT8 pretraining method for transformers. Specifically, to improve training efficiency, we propose using \textbf{INT8 data flow}. As shown in Fig.~\ref{Fig: INT8 training pipeline}, the INT8 data flow simply refers to the utilization of 8-bit integers for data movement among operators.  Compared to the FP16 data flow, the INT8 data flow is 2x faster in theory. In particular, the INT8 data flow considerably enhances the speed of memory-constrained operators, including Layernorm and GELU.

In addition to INT8 flow, we propose \textbf{per-block quantization} that is specialized for transformer pretraining. 
On one hand, compared to per-tensor or per-token quantization \cite{INT8CLIP},  our per-block quantization better preserves the accuracy of pretrained transformers. On the other hand,  per-block quantization brings practical training speedup on tensor cores compared to per-channel quantization.  Furthermore, our method is applicable to a wide range of computing platforms supporting INT8 matrix multiplications (MMs).

We validate our INT8 FQT method for transformers across a diverse range of tasks, including machine translation, image classification, and generative model pretraining. Jetfire consistently attains comparable accuracy with the FP16 training baseline and has superior accuracy compared with the existing works of INT8 training~\cite{INT8CLIP}. On NVIDIA RTX 4090 GPUs, our custom linear and non-linear operators achieve speedups of 1.4x and 1.8x, respectively, compared to the FP16 baselines. Besides, our Jetfire achieves a speedup of 1.42x for a single transformer block and 1.49x memory reduction compared with the FP16 baseline.

\section{Related Work}\label{Sec: Related}
\paragraph{Post-Training Quantization and Quantization-Aware Training}
Post-Training Quantization (PTQ) and Quantization-Aware Training (QAT) aim to find a good low-precision representation for a full-precision model. 
Post-Training Quantization (PTQ) \cite{chee2023quip, xiao2023smoothquant, LLMINT8, kim2023squeezellm, lin2023awq, kim2021ibert, jacob2018INTquantization, liu2021QVIT} converts the pre-trained model's weights to lower-bit representations directly.
Quantization-Aware Training (QAT)~\cite{HAWQ, HAWQV2, Q-BERT, Ternarybert, Binarybert, MKQBERT, esser2019learned} involves retraining the model to adapt its weights and regain accuracy after the quantization process.

\begin{table}[t]
\caption{Comparison with related works. SB refers to SwitchBack~\cite{INT8CLIP}, TE refers to TransformerEngine~\cite{nvidiaTransformerEngine}.}
\label{Table: Related works}
\begin{center}
\begin{footnotesize}
\begin{sc}
\resizebox{0.98\linewidth}{!}{
\begin{tabular}{c|ccccc}
\toprule
  Support & Jetfire (Ours) & SB & TE & FP8-LM & DAQ \\
\midrule

Transformers & $\checkmark$ & $\checkmark$ & $\checkmark$ & $\checkmark$ & $\times$ \\
INT8 quantization & $\checkmark$ & $\checkmark$ & $\times$ & $\times$  & $\checkmark$ \\
8-bit gradient & $\checkmark$ & $\times$ & $\times$ & $\checkmark$ & $\times$ \\
8-bit data flow & $\checkmark$ & $\times$ & $\times$ & $\times$ & $\times$\\
\bottomrule 
\end{tabular}}
\end{sc}
\end{footnotesize}
\end{center}
\vspace{-0.7cm}
\end{table}
\paragraph{Fully Quantized Training}\label{Subsec: FQT}
Fully Quantized Training (FQT)\cite{2018FP8, 2018INT8, INT4Train,FP8graphcore, INT8CLIP, INT8CNN, INT8Distributed, AMP, nvidiaTransformerEngine} has been introduced as a technique to accelerate the training process of neural networks. FQT requires quantizing both the forward propagation and backward propagation to actually accelerate the whole training process. Nowadays 16-bit quantization has been commonly employed with float16 and bfloat16 data formats in training. It introduces loss scaling to prevent underflow and overflow problem. 

For INT8 training, the majority of the work focuses on quantization of CNNs~\cite{INT8CNN, INT8Distributed, zhou2021octo}. \textbf{SwitchBack}~\cite{INT8CLIP} introduces per-token quantization and successfully applies INT8 training to CLIP models for the first time, but still leaves the calculation of weight gradient in FP. 
To be more specific, in the forward process of $\Yv = \Xv\Wv^\top$, they apply per-token quantization for $\Xv$ and per-channel quantization for $\Wv^\top$. In the backward process, for $\nabla\Xv = \nabla\Yv\Wv$, they apply per-token quantization for $\nabla\Yv$ and per-channel quantization for $\Wv$, and leave the calculation of $\nabla\Wv = \nabla\Yv^\top \Xv$ in full precision. For LLM pre-training, per-token quantization still results in significant accuracy loss due to

With the introduction of the Hopper architecture, FP8 training has also gained attention. TransformerEngine~\cite{nvidiaTransformerEngine} incorporates per-layer scaling to reduce quantization errors and proposes using E4M3 during forward and E5M2 during backward passes to adapt. \cite{FP8graphcore} explores adjusting per-tensor scaling biases to improve accuracy, while \cite{FP8LM} investigates further quantizing optimizer states and the weight's master copy to FP8. However, these methods rely on GPUs with the Hopper architecture and cannot be applied to a wider range of GPUs.

As summarized in Table.~\ref{Table: Related works}, our method supports INT8 quantization, 8-bit gradient, and 8-bit data flow at the same time, compared to other FQT methods.



\section{INT8 Data Flow}\label{Sec: Preliminary}

In this section, we introduce our approach for INT8 training with INT8 data flow. We begin by defining the concept of Fully Quantized Training (FQT).

\subsection{Fully Quantized Training}
Consider a network consisting of linear and nonlinear layers. In the forward pass, these layers can be represented as $\Yv = \Fv(\Xv, \Wv)$, where $\Xv$ is the activation, $\Wv$ is the weight, and $\Yv$ is the output, also the next layer's activation. 
In the backward pass, each layer takes the gradient $\nabla_{\Yv}$, $\Xv$, and $\Wv$ as inputs and computes the activation gradient and weight gradient by $\nabla_{\Xv}, \nabla_{\Wv}=\dv\Fv(\nabla_{\Yv}, \Xv, \Wv)$.

Quantization accelerates training by utilizing low-precision computing units on hardware. One notable example is matrix multiplication (MM) in the form of $\Yv=\Xv\Wv^\top$. When both input matrices are in low-precision format, the MM can have 2x theoretical flops relative to an MM with full-precision inputs, where in this paper we assume that the full-precision format is FP16 and the low-precision format is INT8. Most FQT methods utilize such low-precision MM by a \emph{quantize-compute-dequantize (QCD) approach}: (1) temporarily converting FP16 input matrices to INT8 with a \emph{quantizer} $Q(\cdot)$; (2) perform the INT8 MM to get an INT32 output; and (3) convert the output matrix back to FP16 with a \emph{dequantizer} $Q^{-1}(\cdot)$. With QCD, a MM operator can be formulized as $\Yv=\text{QCD-MM}(\Xv, \Wv)=Q^{-1}(Q(\Xv)Q(\Wv^\top))$. As the QCD-MM operator has identical interface to FP16 MMs (i.e., both input and output are still in FP16), we can accelerate training by simply replacing all FP16 MM operators with QCD MMs.

However, QCD only reduces the \emph{computing precision} to INT8, while leaving the \emph{data flow precision} in FP16. That is, MMs are performed under INT8, but the input, output, and data transferred between layers are still in FP16, as illustrated in Fig.~\ref{Fig: INT8 training pipeline}. The practical speedup of  QCD  is limited by the \emph{memory bandwidth}. Modern GPUs have excessive computational power, while the GPU memory bandwidth is scarce. An algorithm must have a high arithmetic intensity (i.e., ratio of computing operations to memory accesses) to run fast on GPUs. Unfortunately, the QCD approach's arithmetic intensity is low: the computation is cut by half, but the memory access is not reduced as much, since data are still represented in FP16. More specifically, QCD has three drawbacks:

\noindent 1. Frequent quantization and dequantization operations incur additional memory access overhead.\\
2. Nonlinear operators cannot be accelerated.\\
3. GPU memory consumption and communication costs remain high.

\subsection{FQT with INT8 Data Flow}

To address these challenges, we directly utilize INT8 \emph{data flow}  throughout the network. That is, we employ the INT8 data format for activations, weights, and gradients, and both our linear and non-linear operators directly take INT8 matrices as inputs and get INT8 tensors as outputs. 

To achieve this, we directly represent activation, weight, and gradient tensors in a custom INT8 format defined in Sec.~\ref{Sec: Method}. 
Then, we redesign and implement all operators used in transformer training, including linear operators (Sec.~\ref{sec: Linear Layer}) and nonlinear operators (Sec.~\ref{Sec: Non Linear Layer}), allowing them to directly use our custom INT8 format as inputs/outputs rather than FP16. The custom INT8 format is carefully designed to ensure that the operators can be implemented efficiently on GPUs, while maintaining accuracy. Such INT8 data flow is compared with QCD in Fig.~\ref{Fig: INT8 training pipeline}.

With the INT8 data flow, we reduced the amount of memory access in the training algorithm, resulting in better efficiency. In a nutshell, our operators read/write INT8 data from global memory in a block-wise fashion, and perform the quantize/dequantize/compute operations on chip within shared memory and registers. In this way, both computation and memory access can be reduced by half, and the arithmetic intensity remains high.
A direct consequence is that, our method can accelerate nonlinear operators, since their memory access is also cut by half.
Finally, as the data are stored in INT8 format, the activation memory consumption  and amount of communication (tensor / pipeline parallelism) can be also cut by half, effectively avoiding memory capacity and communication bottlenecks.

\section{Per Block Quantization}\label{Sec: Method}

\begin{figure}
    \centering
    \subfloat{\label{Fig: QTensor Activation Distribution}{\includegraphics[width=0.45\linewidth]{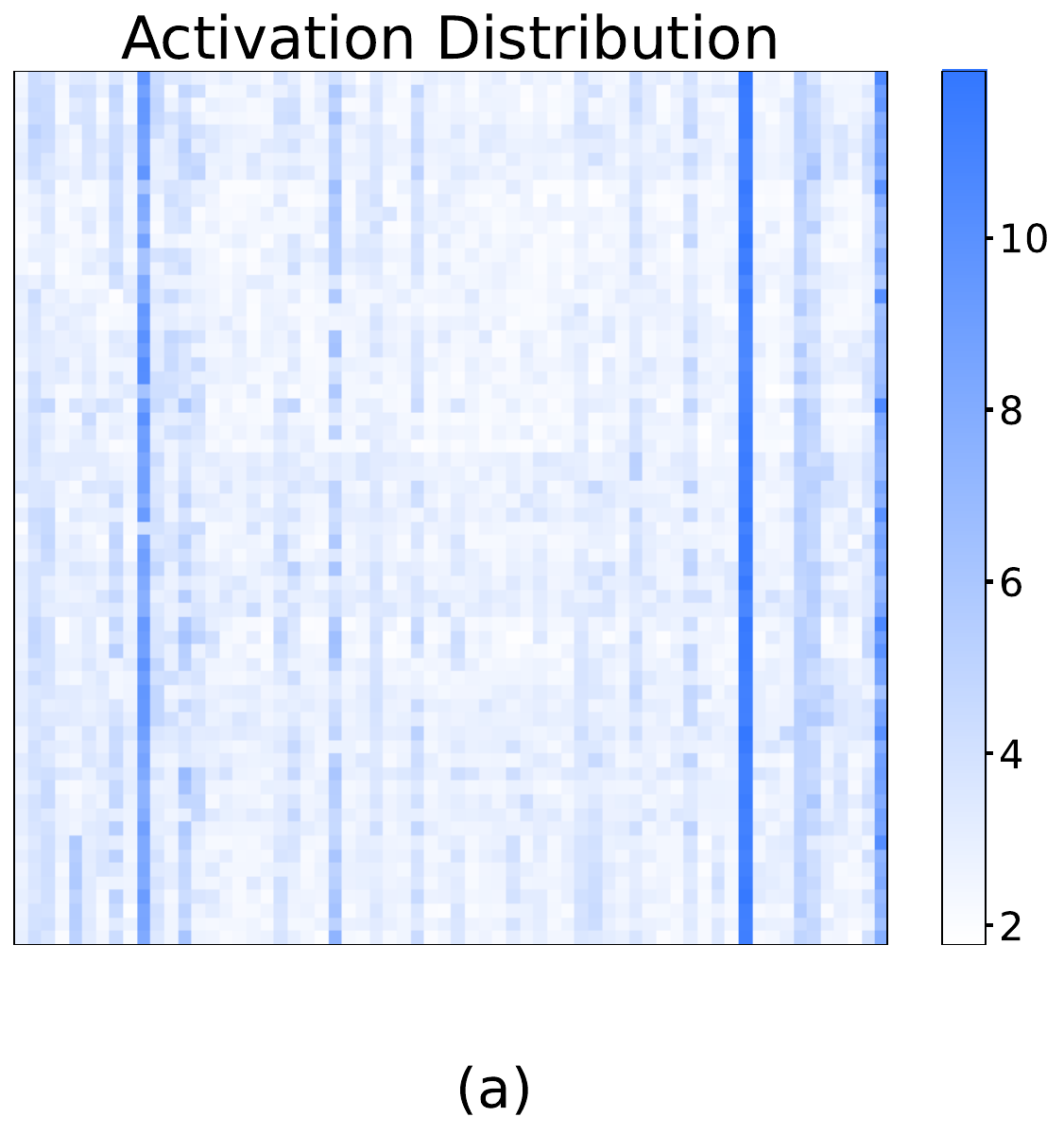}}}
    \subfloat{\label{Fig: GELU memory bound}{\includegraphics[width=0.47\linewidth]{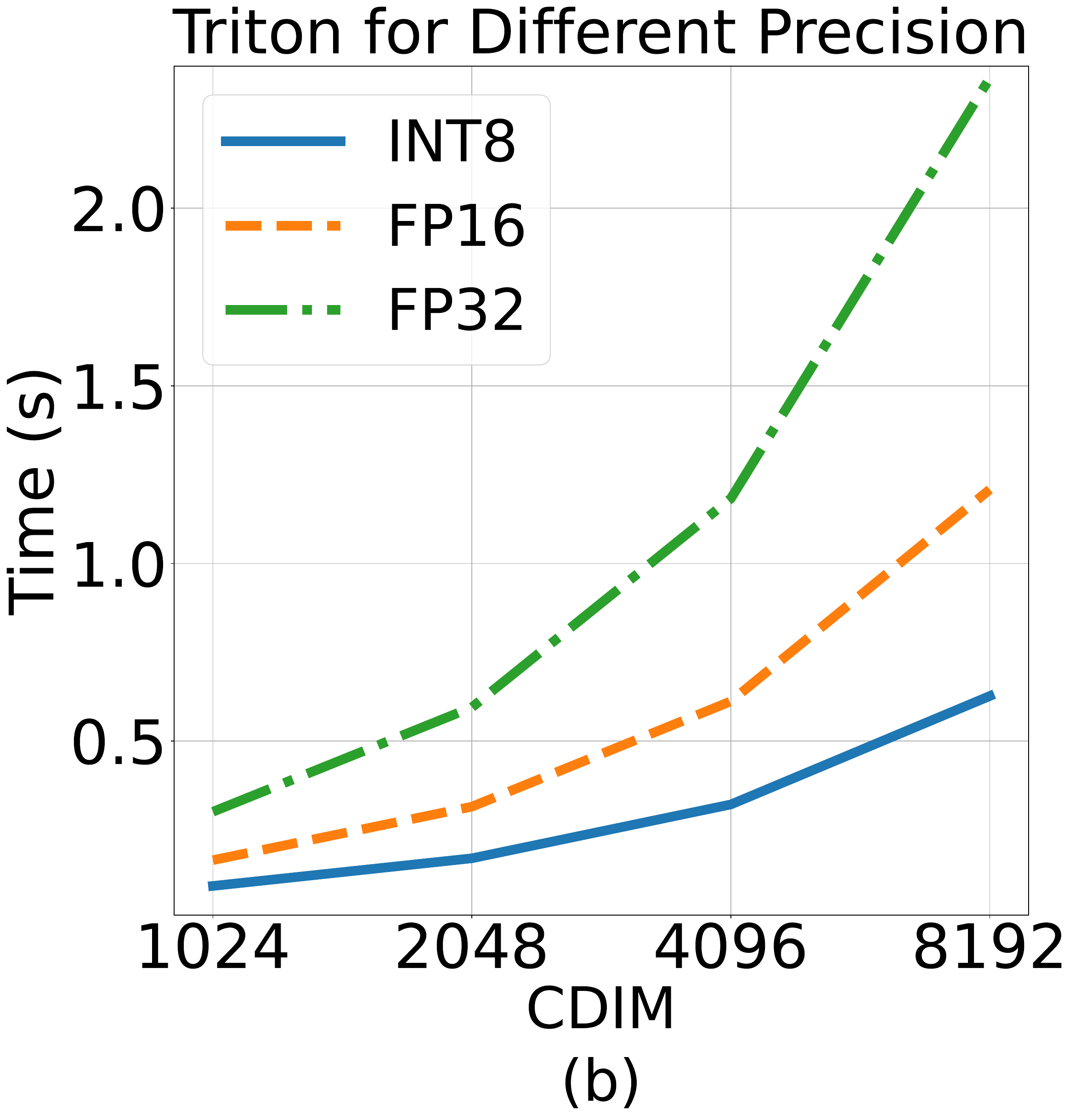}}}
    \vspace{-0.2cm}
    \caption{(a) Channel-wise outliers in activation distribution. (b) Non-linear operator is memory-bounded.}
\vspace{-0.5cm}
\end{figure}



In this section, we introduce our INT8 numerical format. Typically, we can approximate an FP16 matrix with an INT8 matrix $\Xv^{\text{INT8}}$ and a FP16 scale factor $\Sv_{\Xv}^{\text{FP16}}$, that is
$\Xv^{\text{INT8}}, \Sv_{\Xv}^{\text{FP16}} = Q(\Xv^{\text{FP16}})$.
Depending on the shape of the scale factor, there are different quantization methods, including per-tensor quantization, per-token quantization, and per-channel quantization. 
The INT8 numerical format must accurately support the following three MMs of a linear layer in forward and back propagation:
\begin{align*}
\Yv=\Xv\Wv^{\top},~~~\nabla_{\Xv}=\nabla_{\Yv} \Wv,~~~\nabla_{\Wv}=\nabla_{\Yv}^\top  \Xv.
\end{align*}

Researchers have observed that activations in transformers are difficult to quantize~\cite{LLMINT8, xiao2023smoothquant} due to the presence of \textbf{channel-wise outliers}. We visualize this problem in Fig.~\ref{Fig: QTensor Activation Distribution}. 
Per-token quantization assigns different scale factors for different tokens and often results in large quantization errors since outliers appear channel-wise. On the other hand, per-channel quantization assigns different scale factors for different channels and has relatively lower quantization errors, as shown in Sec.~\ref{Subsec: Quantization Error}. 
In addition, gradient outliers also appear along the token axis~\cite{chen2020statquant,INT4Train},  which poses challenges for computing the weight gradient $\nabla_{\Wv}=\nabla_{\Yv}^\top  \Xv$. In this case, per-token quantization should be applied to the output gradient $\nabla_{\Yv}$ to avoid large quantization error. 

However, applying per-channel quantization for forward propagation or applying per-token quantization for computing weight gradients both pose challenges in practical hardware implementations. For a MM in the form $\Cv=\Av\Bv$, we call the 0th axis of $\Av$ and the 1st axis of $\Bv$ to be outer axes, as $\Cv$ has them; the other two axes are inner axes. 
INT8 MMs are performed with tensor core WMMA (Warp Matrix Multiply-Accumulate) operations~\cite{markidis2018Tensorcore}, and scaling can only be performed at the outer axis of MM if we want to utilize tensor core. As a compromise,~\cite{INT8CLIP} only use per-token quantization for forward propagation, sacrificing accuracy; and fall back to FP16 when computing weight gradients, sacrificing speed.



\begin{figure}[t]
\centering
\subfloat{{\includegraphics[width=0.9\linewidth]{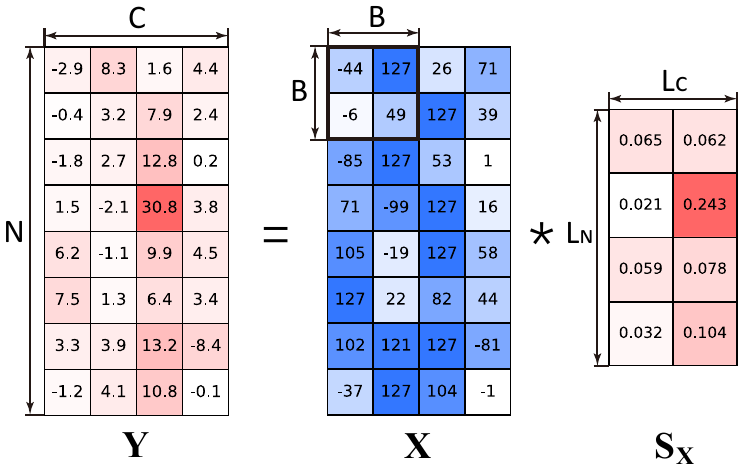}}}
\vspace{-0.2cm}
\caption{Visualization of the per-block quantization methodology. When the original tensor has some outliers, our method can restrict its effect to a $B\times B$ block.}
\label{Fig: quantization method visualization}
\vspace{-0.3cm}
\end{figure}

We propose \emph{per-block quantization} to achieve computational efficiency and preserve accuracy at the same time. For a matrix $\Xv \in \Rb^{N\times C}$, we partition $\Xv$ into blocks $\Xv_{ij}\in \Rb^{B \times B}$ along both row axis and column axis, where $B$ is quantization block size, $i, j$ is the index of quantization block along the token and channel axis. We assign a scale factor $\sv_{ij}$ for each block $\Xv_{ij}$ that corresponds to the maximum absolute value in the block. The method can be formulated as:
\begin{align}\label{Eq: per-block quantization}
\small
    Q(\Xv_{ij}) = \left\lceil\frac{\Xv_{ij}}{\sv_{ij}}\right\rfloor, ~~
    Q^{-1}(\Xv_{ij}^{\text{INT8}}, \sv_{ij}) = \Xv_{ij}^{\text{INT8}}\sv_{ij},
\end{align}
where $\lceil\cdot\rfloor$ is the round operator. We visualize this method in Fig.~\ref{Fig: quantization method visualization} for better understanding. Since our per-block quantization method partitions along the inner axis, it restricts the impact of an outlier channel/token within a block. Therefore the quantization error is controlled. We will demonstrate in the next section that per-block quantization can be also efficiently implemented on GPUs.

\vspace{-0.2cm}
\section{Linear Layer Operator}\label{sec: Linear Layer}
In this section, we mainly discuss how our per-block quantization method should be applied to linear layers. 
We highlight that our linear operator adopts INT8 data flow, that takes INT8 as input and produces INT8 as output.

\begin{figure*}[t]
\centering
\subfloat{\label{Fig: Quantized Linear Layer Algorithm}{\includegraphics[width=0.9\linewidth]{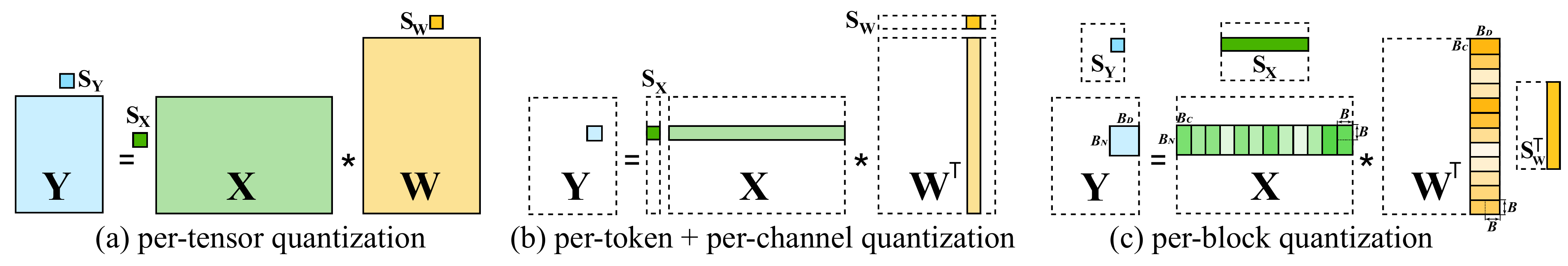}}}
\caption{Different quantization methods for linear layer.}
\label{fig:2}
\vspace{-0.6cm}
\end{figure*}

\subsection{Notations}

We consider the CUDA implementation of the following MM as an example in this section:
\begin{align}\label{Eq: Forward MM}
\resizebox{0.9\linewidth}{!}{$
\Yv=\Xv\Wv^\top,
\text{where}~
\Xv\in\Rb^{N\times C}, \Wv\in\Rb^{D\times C},
\Yv\in\Rb^{N\times D},
$}
\end{align}
which dimensions are represented as $N\times C\times D$.

In our MM operator, each input and output matrix is represented in per-block INT8 format: a INT8 matrix and a FP16 scale matrix, as defined in Sec.~\ref{Sec: Method}. In this case, we have INT8 input denoted as $\Xv$ and $\Wv$, and we have scale factors denoted as $\Sv_{\Xv}\in\Rb^{L_N \times L_C}, \Sv_{\Wv}\in\Rb^{L_D \times L_C}, $ where $L_N=\tfrac{N}{B}, L_C=\tfrac{C}{B}, L_D=\tfrac{D}{B}$ is the number of quantization blocks along every axis, and $B$ is the quantization block size in Eq.~(\ref{Eq: per-block quantization}).  We utilize tensor cores to perform INT8 WMMA. For a single INT8 WMMA instruction, the inputs are two INT8 matrices of shape $16\times 16$ and the output is an INT32 matrix of shape $16\times 16$. 

\subsection{3-Level Tiling of MM}
An efficient MM implementation must organize the computation into blocks (``tiling'') based on the GPU architecture. We tile the computation in 3 levels. The block dimensions are listed in Table~\ref{Table: Variable meaning}.
\paragraph{CUDA block level}
When implementing our MM operator in CUDA, we first parallelize the computation along the $N$ and $D$ axis. 
Every time we only calculate a submatrix $\Bv_{\Yv} \in \Rb^{B_N\times B_D}$ of the output matrix $\Yv$. We further divide $C$ into small segments of size $B_C$, and accumulate along this axis. 
The CUDA block size $B_N \times B_C \times B_D$ is architecture specific. Depending on shared memory capacity and number of threads, typical values are 32, 64, 128, or 256. 
We define $T_N = \left\lceil N/B_N \right\rceil, T_C = \left\lceil C/B_C \right\rceil, T_D = \left\lceil D/B_D \right\rceil$ to be the number of blocks along each axis of the MM. 

For every iteration, we load submatrix $\Xv_{ik} \in \Rb^{B_N\times B_C}$ and $\Wv_{jk} \in \Rb^{B_D\times B_C}$ from global memory to shared memory and compute the output submatrix $\Yv_{ij} \in \Rb^{B_N\times B_D}$, where $1 \leq i \leq T_N, 1 \leq j \leq T_D, 1 \leq k \leq T_C$ are the CUDA block index along the $N, D, C$ axis. 

\paragraph{Quantization block level}
We set $B_N$ and $B_D$ to be  multiples of the quantization block size $B$, and set $B_C=B$. In this case, $\Xv_{ik}$ consists of $R_N = \left\lceil B_N/B \right\rceil$ quantization blocks along its 0th-axis, and we use $\Xv_{ik,p}$ to denote the $p$-th quantization block. Similarly, $\Wv_{jk}$ consists of $R_D = \left\lceil B_D/B \right\rceil$ quantization blocks along its 0th-axis with $\Wv_{jk,q}$ as the $q$-th block. 

We use two nested \emph{for loops} to iterate over $R_N$ and $R_D$, load $\Xv_{ik, p} \in \Rb^{B \times B}$ and $\Wv_{jk, q} \in \Rb^{B \times B}$ from shared memory to register and performing INT8 WMMA separately to get INT32 output $\Yv_{ij, pq} \in \Rb^{B \times B}$, where $1 \leq p \leq R_N, 1 \leq q \leq R_D$ is the quantization block index along $R_N, R_D$ axis. 

\paragraph{WMMA operation level}
Within the computation of single quantization blocks, we utilize the INT8 WMMA instruction for computation on register. Therefore, when we set $B=32$ as an example, we need to perform $2^3=8$ WMMA instructions to complete the computation, since a single WMMA instruction can only compute $16\times 16\times 16$ MM.

In summary, we divide the implementation of the MM operator into three levels. First, at the CUDA block level, we divide the operator into sizes of $B_N \times B_C \times B_D$ for computation. Then, at the quantization block level, we further divide each CUDA block into sizes of $B \times B \times B$. Finally, at the WMMA operation level, we divide the computation of each quantization block based on the dimensions of the WMMA operation. 

\subsection{Quantize and Dequantize}
We now discuss how to integrate the quantize and dequantize operators in our algorithm. Since different quantization blocks have different scale factors, after every INT8 WMMA operation, we need to dequantize the INT32 output into FP32 and accumulate in FP32. By applying the same index notation as the previous section, we have
\begin{align*}
\Yv_{ij, pq}^{\text{INT32}} = \Xv_{ik, p}\Wv_{jk, q}^\top,
~~\Yv_{ij, pq}^{\text{FP32}} = \sum_{k=1}^{T_C} \sv_{\Xv_{ik, p}}^{\text{FP16}}\Yv_{ij, pq}^{\text{INT32}}{\sv_{\Wv_{jk, q}}^{\text{FP16}}},
\end{align*}
where $\sv_{\Xv}, \sv_{\Wv}$ is scale factor and both $\Yv$s are accumulators. 

After the calculation of $\Yv_{ij, pq}^{\text{FP32}}$, we quantize it to get a INT8 submatrix $\Yv_{ij, pq}^{\text{INT8}}$ and a scale factor $\sv_{\Yv_{ij, pq}}$.

We formalize our algorithm in Algorithm~\ref{Alg: INT8 Linear Layer}. In the algorithm, we have omitted the details of the quantization block level and WMMA operation level for simplicity. We highlight the overhead introduced by our method in red.
We further compare it with per-tensor quantization MM~\cite{2018INT8} and per-token quantization MM~\cite{INT8CLIP} in Fig.~\ref{Fig: Quantized Linear Layer Algorithm}.

Our algorithm accurately quantizes channel-wise outliers while introducing only a small amount of overhead for dequantize and quantize operations. We calculate the overhead within the computation of a submatrix $\Yv_{ij}$ and compare our method with basic INT8 MM and SwitchBack. Results are reported in Table ~\ref{Table: MM operation complexity}. 
The time complexity of MM is $O(B_N * C * B_D)$. while our method's overhead time complexity is $O(B_N * T_C * B_D) + (BN + B_D)C$. Since $\frac{C}{T_C} = B_C$ is typically set to 32 or 64 and $B_N, B_D$ is 128 or 256, the overhead is negligible.

\begin{table}[t]
\vspace{-0.2cm}
\centering
\caption{Meaning of Key Constants.}
\label{Table: Variable meaning}
\begin{footnotesize}
\begin{tabular}{@{}ll@{}}
\hline
$B_N/B_C/B_D$ & CUDA block size in MM \\ 
              \hline
$T_N/T_C/T_D$ & Number of CUDA blocks along each axis \\
\hline
$B$ & Quantization block size\\ 
              \hline
$R_N/R_C/R_D$ & Number of quantization blocks \\
              & in a CUDA block along each axis 
              \\ \hline
\end{tabular}
\end{footnotesize}
\vspace{-0.5cm}
\end{table}

\begin{table}[]
\centering
\caption{Time complexity of different operations in MM.}
\label{Table: MM operation complexity}
\begin{footnotesize}
\begin{sc}
\resizebox{\linewidth}{!}{
\begin{tabular}{cccc}
\toprule
 ~ &  \multicolumn{3}{c}{Method}    \\ 
\cmidrule(r){2-4}
Operation & Basic INT8 & SwitchBack & Ours \\ 
\midrule
MM & $B_NB_DC$ & $B_NB_DC$ & $B_NB_DC$ \\
\midrule
16-bit Load/Store & $(B_N+B_D)C + B_NB_D$ & $\frac{B_NB_C}{T_D} + \frac{B_DB_C}{T_N} + B_NB_D$ & -  \\
\midrule
8-bit Load/Store & - & $(B_N+B_D)C$ & $(B_N+B_D)C + B_NB_D$  \\
\midrule
Dequantize & - & $B_NB_D$   & $B_NB_DT_C$  \\
\midrule
Quantize & - & $\frac{B_NB_C}{T_D} + \frac{B_DB_C}{T_N}$ & $B_NB_D$  \\
\bottomrule
\end{tabular}
}
\end{sc}
\end{footnotesize}
\vspace{-0.5cm}
\end{table}




\begin{algorithm}[t]
\small
\caption{INT8 Linear Layer}\label{Alg: INT8 Linear Layer}
\begin{algorithmic}[1]
\REQUIRE INT8 Matrices $\Xv \in \Rb^{N \times C}, \Wv \in \Rb^{D \times C}$, FP16 scale matrices $\Sv_{\Xv} \in \Rb^{L_N \times L_C}, \Sv_{\Wv} \in \Rb^{L_D \times L_C}$, CUDA Block size $B_N \times B_C \times B_D$

\STATE Define $T_N = \left\lceil \frac{N}{B_N} \right\rceil$, $T_C = \left\lceil \frac{C}{B_C} \right\rceil$, $T_D = \left\lceil \frac{D}{B_D} \right\rceil$

\STATE Define $R_N = \left\lceil \frac{B_N}{B} \right\rceil, R_C = \left\lceil \frac{B_C}{B} \right\rceil R_D = \left\lceil \frac{B_D}{B} \right\rceil$
\FOR{1 $\le$ i $\le T_N$}
    \FOR{1 $\le$ j $\le T_D$}
    \STATE Initialize accumulator \textcolor{red}{$\Yv_{ij}^{\textbf{FP32}}$}, $\Yv_{ij}^{\textbf{INT32}}$
        \FOR{1 $\le$ k $\le T_C$}
        \STATE Load INT8 Block $\Xv_{ik} \in \Rb^{B_N \times B_C}$ and scale factor $\Sv_{\Xv_{ik}} \in \Rb^{R_N \times R_C}$
        \STATE Load INT8 Block $\Wv_{jk}^\top \in \Rb^{B_C \times B_D}$ and scale factor $\Sv_{\Wv_{jk}^\top} \in \Rb^{R_C \times R_D}$
        \STATE On chip, compute INT8 Matmul: $\Yv_{ij}^{\textbf{INT32}} = \Xv_{ik}\Wv_{jk}^\top$
        \STATE On chip, \textcolor{red}{dequantize to FP32} and accumulate: \\$\Yv_{ij}^{\textbf{FP32}} \leftarrow \Yv_{ij}^{\textbf{FP32}} + $ \textcolor{red}{$\space\Sv_{\Xv_{ik}}\Yv_{ij}^{\textbf{INT32}}\Sv_{\Wv_{jk}^\top}$}
        \ENDFOR

        \STATE On Chip, \textcolor{red}{quantize the output $\Yv_{ij}^{\textbf{FP32}}$} to get $\Yv_{ij}^{\textbf{INT8}}\in \Rb^{B_N \times B_D}$ and scale $\Sv_{ij} \in \Rb^{R_N \times R_D}$
        \STATE Save $\Yv_{ij}^{\textbf{INT8}}$ and $\Sv_{\Yv_{ij}}$ to global memory.
    \ENDFOR
\ENDFOR
\end{algorithmic}
\end{algorithm}

\section{Non-Linear Operator}\label{Sec: Non Linear Layer}
In this section, we mainly discuss how our per-block quantization method should be applied to non-linear layers. By reducing the precision of the input and output to INT8, we can achieve acceleration for these operators as well.

\subsection{Non-Linear Operators are Memory-Bounded}
We have observed that non-linear operators are memory-bounded, which means that the speed of these operators is primarily limited by memory bandwidth, rather than by computation. We validate this by manipulating the data format (INT8, FP16, FP32) for global memory read/write operations in the GELU operator, while internally converting them to FP32 for computation. Fig.~\ref{Fig: GELU memory bound} illustrates that even computations are kept in FP32, simply reducing the read/write precision can already obtain near-linear speedup. As our method reduces the data flow precision from FP16 to INT8, we anticipate $\sim$2x speedup for all nonlinear operators. In contrast, QCD cannot accelerate nonlinear operators.


\subsection{Triton Implementation}
Based on the observations above, our main idea is to load/write in INT8 and leave all calculations within the shared memory through kernel fusion. Specifically, after loading the INT8 input into shared memory, we dequantize it to FP32 and apply the non-linear operators, then quantize the FP32 output back to INT8 format before writing the data into global memory.

We primarily focus on non-linear operators like GELU~\cite{GELU}, 
LayerNorm~\cite{Layernorm}, Dropout~\cite{dropout}, and Add~\cite{resnet}, and implement them with Triton~\cite{tillet2019triton}. 

We define $f$ to be the element-wise operator, $\Xv, \Yv \in \Rb^{N\times C}$ to be the INT8 input and output, $\Sv_{\Xv}, \Sv_{\Yv} \in \Rb^{L_N \times L_C}$ are scale factors, where $L_N = \frac{N}{B}, L_C = \frac{C}{B}$ are number of quantization blocks along each axis and $B$ is the quantization block size. Similar to CUDA, we also do tiling to parallelize the computation. For a single block (whose shape is defined as Triton Block Size) we denote $\Xv_{ij}^{\textbf{INT8}}$ to be the input tensor and $\Sv_{\Xv_{ij}}^{\textbf{FP16}}$ to be the scale. The computation process can be represented as 
\begin{align*}
\begin{small}
\Yv_{ij}^{\textbf{FP32}} = f(Q^{-1}(\Xv_{ij}^{\textbf{INT8}}, \sv_{\Xv_{ij}}^{\textbf{FP16}}));~~~
\Yv_{ij}^{\textbf{INT8}}, \sv_{\Yv_{ij}}^{\textbf{FP16}} = Q(\Yv_{ij}^{\textbf{FP32}}),
\end{small}
\end{align*}
where $Q^{-1}$ and $Q$ is the dequantizer and quantizer, $\Yv_{ij}^{\textbf{INT8}}$ is the output tensor, and $\sv_{\Yv_{ij}}^{\textbf{FP16}}$ is the scale factor. This algorithm can be expressed as Algorithm ~\ref{Alg: Non-linear}, where we omit the quantization block level for simplicity.

\section{Experiments}\label{Sec: Experiments}

\begin{figure}[t]
\centering
\subfloat{{\includegraphics[width=0.95\linewidth]{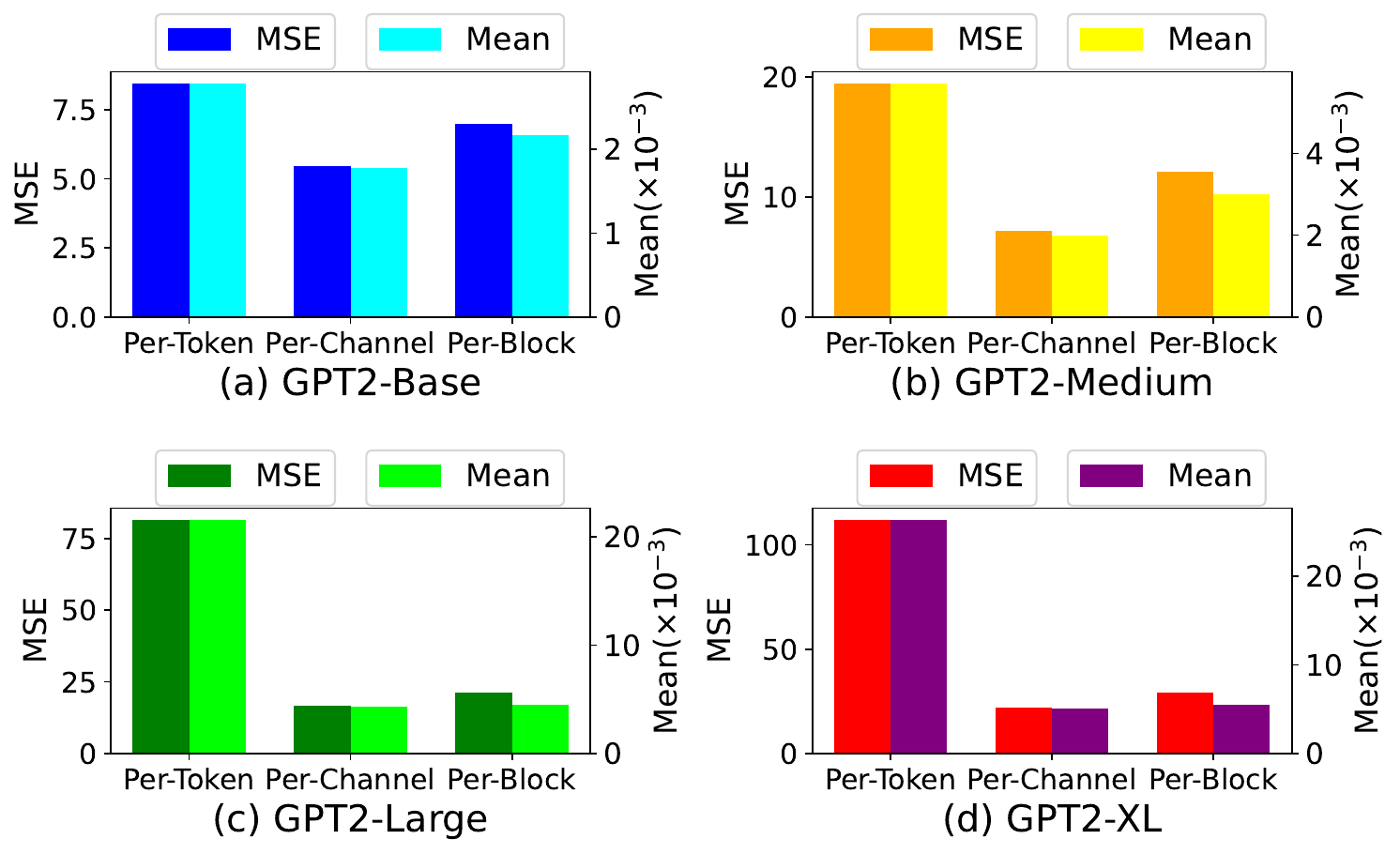}}}
\caption{Quantization error for different quantization methods. Per-Block refers to our Jetfire quantization method.}
\label{Fig: Quantization Error of different methods}
\end{figure}

\begin{figure}[t]
\centering
\subfloat{\label{Fig: GELU Block size Ablation}{\includegraphics[width=0.5\linewidth]{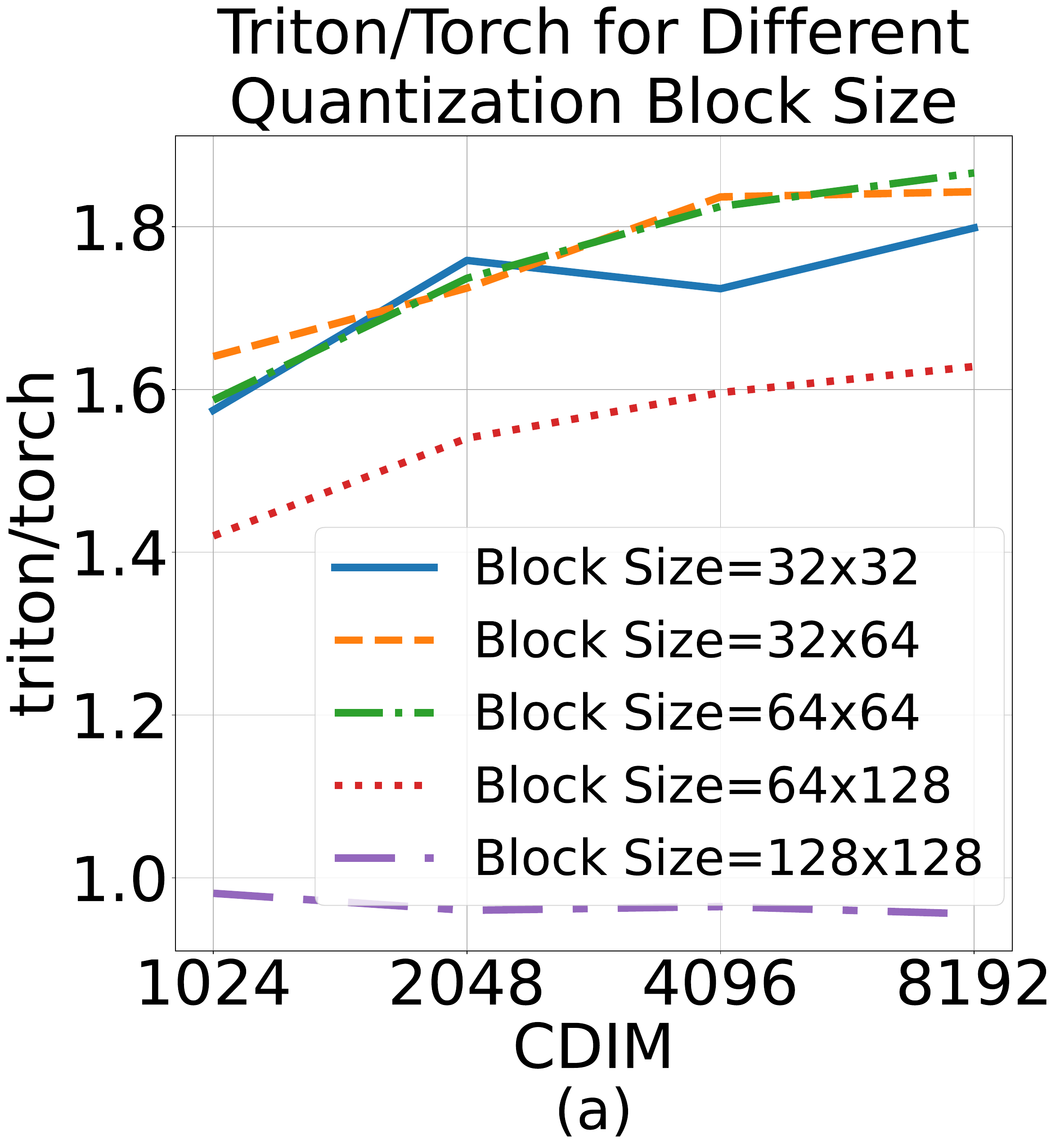}}}
\subfloat{\label{Fig: GEMM CUDA}{\includegraphics[width=0.5\linewidth]{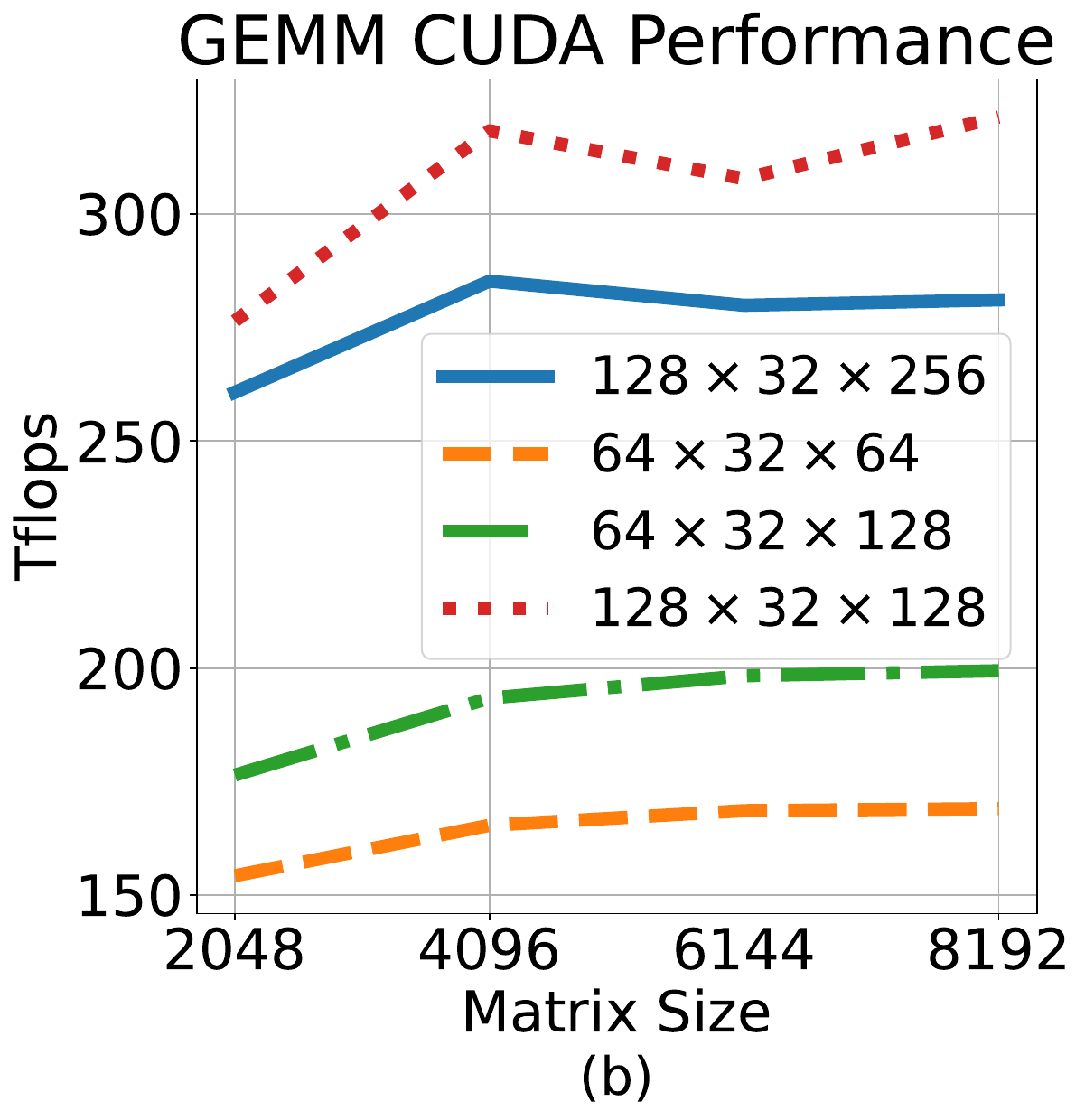}}}
\caption{Speed test of GELU and GEMM operator. (a) Triton kernel speedup with different Triton block sizes.(b) GEMM CUDA kernel speed with different CUDA block sizes.}
\label{Fig: GELU test}
\end{figure}

\begin{figure}[t]
\centering
\subfloat{{\includegraphics[width=0.9\linewidth]{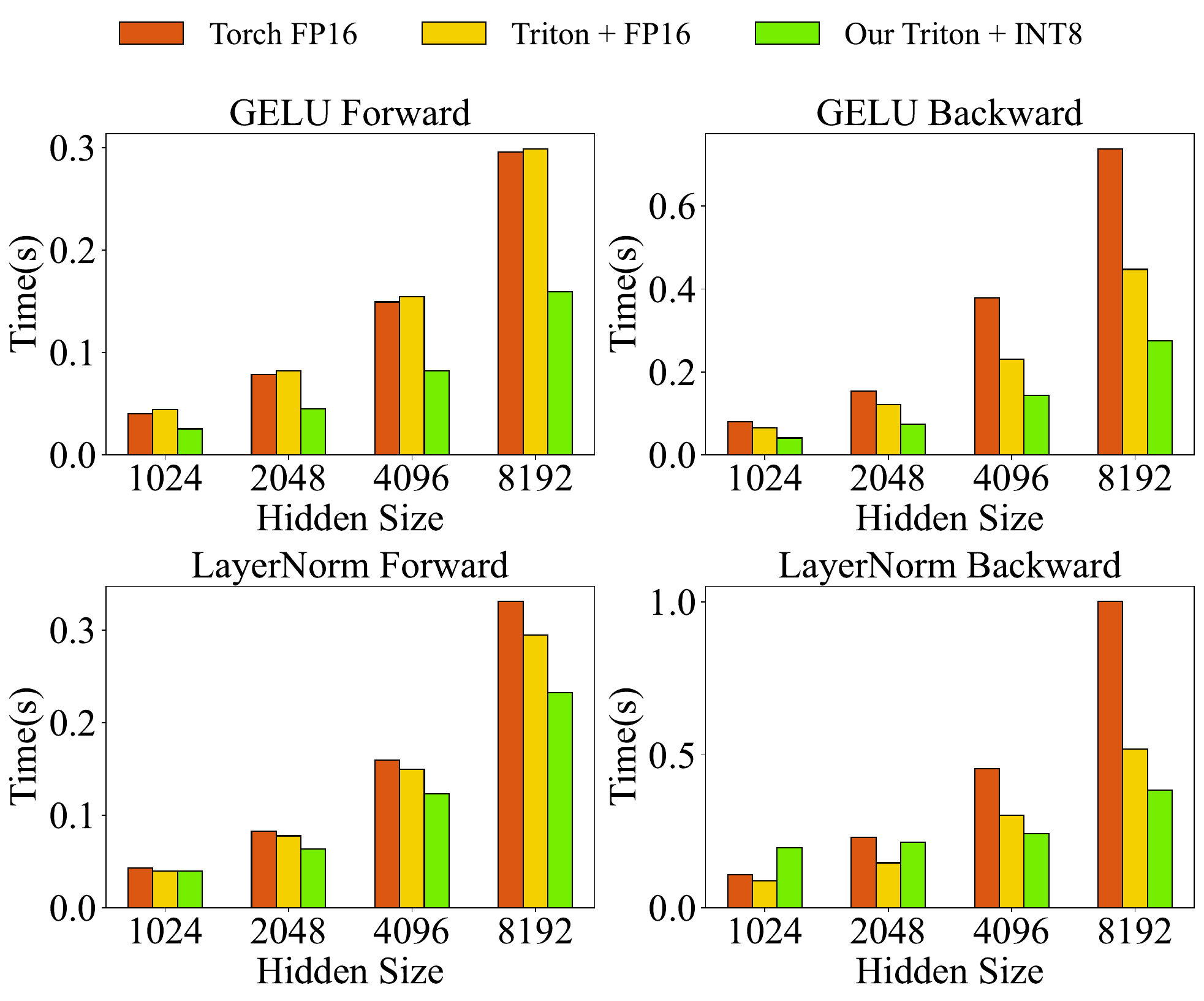}}}
\caption{Speed comparision between our INT8 non-linear operator and pytorch FP16 implementation.}
\label{Fig: GELU and LayerNorm Speed Test}
\end{figure}

\begin{figure}[t]
\centering
\subfloat{{\includegraphics[width=0.85\linewidth]{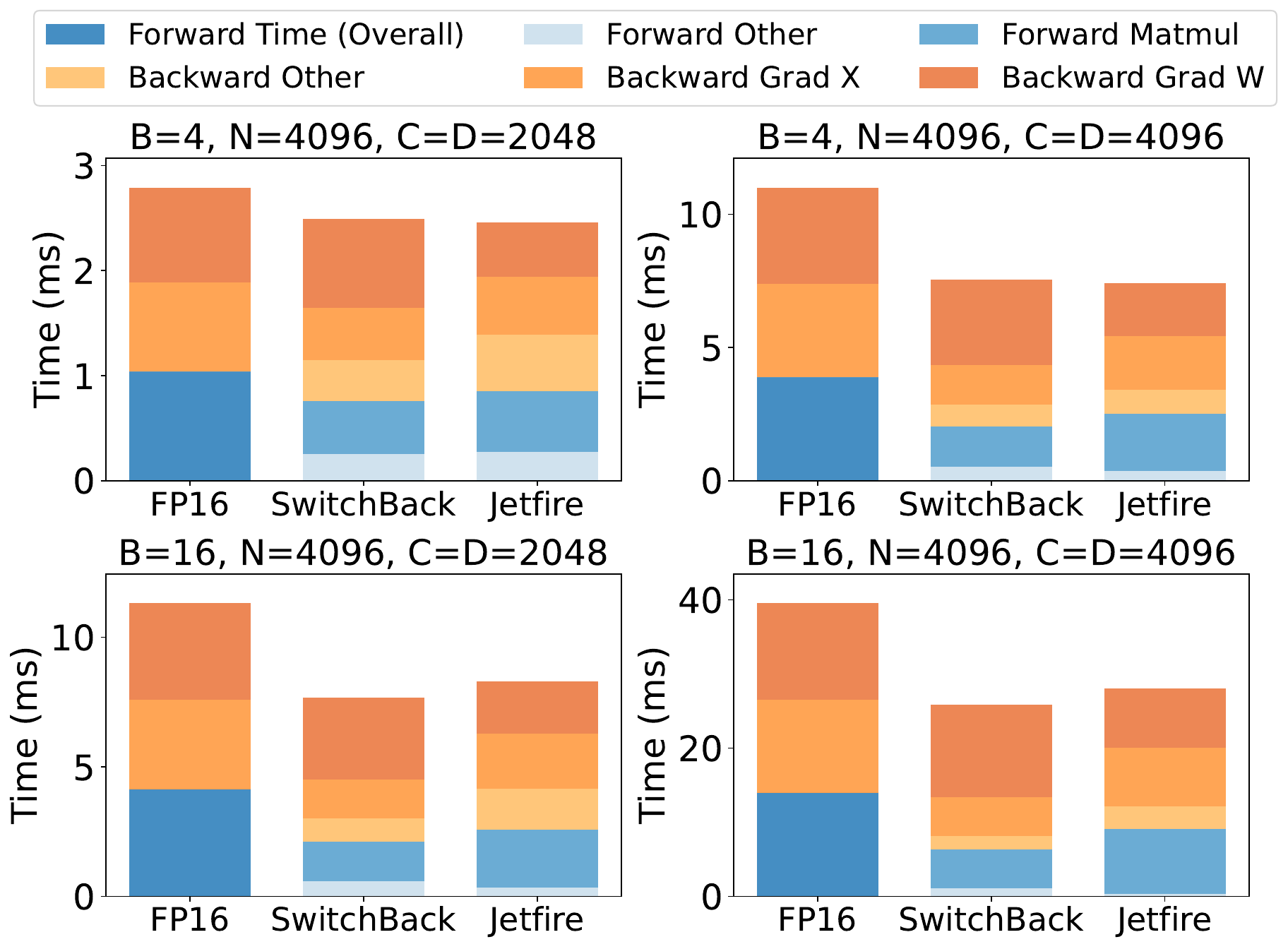}}}
\caption{ Matrix Multiplication Speed test for Different Methods in different settings ($B$=Batch Size, $N$=Sequence Length).}
\label{Fig: Matmul Speed Test}
\end{figure}
\begin{table*}[t]
\vskip -0.1in
\caption{Results on machine translation, deit pretraining, GPT2 pretraining, and GLUE fine-tuning result based on the pretrained model. FP refers to floating-point, SwitchBack refers to per-token quantization. '--' means the model does not  converge.}
\label{Table: Model}
\begin{center}
\begin{scriptsize}
\begin{sc}
\resizebox{0.9\linewidth}{!}{
\begin{tabular}{ccccccc}
\toprule
~ & ~ & ~ & \multicolumn{3}{c}{Baseline} &
Ours
\\
\cmidrule(r){4-6} \cmidrule(r){7-7}
Model & $\#$Params(M) & METRIC & FP & Switchback & Per-Tensor & Jetfire
\\
\midrule
Transformer-Base & 61 & BLEU & 26.49 & 26.46 & 26.04 & \textbf{26.49}
\\
\midrule
Deit-tiny & 5 & \multirow{3}{*}{Top1 Acc} & 64.08 & 63.55 & -- & \textbf{63.95}
\\
Deit-small & 22 & & 73.43 & 72.80 & -- & \textbf{73.31}
\\
Deit-base & 86 & & 75.67 & 75.62 & -- & \textbf{76.03} 
\\
\midrule
GPT2-base & 124 & \multirow{3}{*}{Valid loss} & 2.9074 & 3.0796 & 3.1638 & \textbf{2.8597}
\\
GPT2-medium & 350 & & 2.6612 & 2.9141 & 3.1795 & \textbf{2.4195}
\\
GPT2-large & 774 & & 2.5993 & 3.0512 & 2.9775 & \textbf{2.4696}
\\
\midrule
GPT2-base & 124 & \multirow{3}{*}{GLUE score} & $78.50_{0.45}$ & $78.15_{0.15}$ & $76.33_{0.94}$ & $\boldsymbol{78.18_{0.34}}$
\\
GPT2-medium & 350 & & $82.00_{0.50}$ & $80.03_{0.15}$ & $79.19_{0.22}$ & $\boldsymbol{81.60_{0.26}}$
\\
GPT2-large & 774 & & $83.01_{0.24}$ & $78.74_{0.24}$ & $75.88_{0.35}$ &
$\boldsymbol{82.94_{0.70}}$ \\
\bottomrule
\end{tabular}}
\end{sc}
\end{scriptsize}
\end{center}
\vskip -0.3in
\end{table*}

\begin{table}[t]
    \centering
    \begin{small}
    \begin{sc}
    \begin{tabular}{cccc}
        \toprule
        Model & Swin-tiny & Swin-small & Swin-base \\
        \midrule
        FP      & 77.55 & 80.39 & 80.45 \\
        Jetfire & 77.51 & 80.39 & 80.37 \\
        \midrule
        & ViT-base & ViT-large \\
        \midrule
        FP      & 83.45 & 85.72 & \\
        Jetfire & 83.48 & 85.67 & \\
        \bottomrule
    \end{tabular}
    \end{sc}
    \end{small}
    \caption{Comparison of FP and Jetfire}
    \label{Table: SwinViT}
\end{table}
\subsection{Settings}

We evaluate our INT8 training algorithm Jetfire on a wide variety of tasks including machine translation, image classification, and generative model pretraining. We adopt default architectures, optimizers, and schedulers for all the evaluated models. We adopt the default hyperparameter except for generative model pretraining.

We quantize all of the linear layers in the MLP and attention module and non-linear layers (including GELU, LayerNorm and Dropout) to INT8, and leave multi-head attention in FP16 by employing FlashAttention~\cite{dao2022flashattention}. 
The master copy of the weights is kept in FP32. We quantize linear layers’ weights to INT8 prior to each matmul, but leave layernorm’s weight and bias to floating-point since they are relatively small. 
We compare our method with floating point training baseline (denoted as FP), per-tensor quantization, and SwitchBack~\cite{INT8CLIP}). We do not compare with FP8 training algorithms as they require specialized Hopper architecture GPU to run, making them less accessible. We emphasize that only our method adopts an INT8 data flow and quantizes non-linear layers.

We implement our linear operators with CUDA and implement non-linear operators with Triton. CUDA block size is set to $128\times32\times128$ and Triton block size is set to $64\times 64$. The quantization block size is set to $B=32$.

\subsection{Converged Model Accuracy}
\paragraph{Machine Translation}\label{Subsec: Machine Translation}

We validate our Jetfire's effectiveness on the translation task. We train a Transformer-base model on WMT 14 En-De dataset~\cite{En-De} based on Nvidia's recipe~\footnote{https://github.com/NVIDIA/DeepLearningExamples\\/tree/master/PyTorch/Translation/Transformer}. 
In Table~\ref{Table: Model} we report the BLEU~\cite{Bleu} score result. Our method has no degradation compared with the FP baseline, while the SwitchBack baseline has $0.03\%$ BLEU score degradation, and the per-tensor quantization baseline has $0.4\%$ degradation.

\paragraph{Image Classification - Deit}\label{Subsec: Deit pretraining}
We do pretraining for Deit-Tiny, Deit-Small, and Deit-Base~\cite{touvron2021training} model on ImageNet1K~\cite{deng2009imagenet} for 90 epochs based on facebook research's recipe~\footnote{https://github.com/facebookresearch/deit}. Results are listed on Table~\ref{Table: Model}. In all experiments, Our method has less than $0.1\%$ accuracy degradation compared with the floating-point baseline, and for Deit-base, our method shows $0.4\%$ improvement. For Deit-tiny and Deit-small models, Switchback has over $0.5\%$ accuracy degradation, and per-tensor quantization does not converge. This indicates that our method can accurately quantize channel-wise outliers. Comparison with more baselines~\cite{wang2023GDA, zhao2021DAQ} can be found in Appendix~\ref{Sec: More baselines}

\paragraph{Image Classification - Swin Transformers and ViT}\label{Subsec: SwinViT pretraining}
We do pretraining for Swin-Transformers(Swin-tiny, Swin-small, Swin-base) for 90 epochs and fine-tuned ViT(ViT-base, ViT-large) for 100 epochs without pre-training (MAE includes pretraining and finetuning) on ImageNet1K. We adopt the official training recipe~\footnote{https://github.com/microsoft/Swin-Transformer}\footnote{https://github.com/facebookresearch/mae?tab=readme-ov-file} and default hyperparameters, and only compare with the full precision training baseline. The results are shown in Figure~\ref{Table: SwinViT}. In all of the experiments, our method achieves less than 0.1\% accuracy degradation, which proves the accuracy of our method.
\paragraph{Generative Model Pretraining}\label{Subsec: GPT2}

We evaluate our method by training three GPT2~\cite{radford2019gpt2} models with different sizes: GPT2-base for 300k steps, GPT2-medium for 200k steps, and GPT2-large for 120k steps on the OpenWebText~\cite{openwebtext} dataset based on NanoGPT\footnote{https://github.com/karpathy/nanoGPT.} (Hyperparameters: Learning Rate = $1.5\times 10^{-4}$, Weight Decay = $10^{-1}$). We report the validation loss and the fine-tuning average accuracy on the GLUE~\cite{GLUE} dataset over 3 seeds. The results are shown in Table~\ref{Table: Model}.

We found that SwitchBack resulted in 0.1 valid loss degradation on GPT-base and led to 0.3-0.4 valid degradation on GPT-medium and GPT-large. Our method achieves even lower valid loss compared to the FP baseline, which may be attributed to the regularization effect of quantization. 

For fine-tuning, our method shows less than 0.3\% degradation compared to baseline, while SwitchBack has a degradation of 0.3\% on GPT2-base, 1.8\% on GPT2-medium, and 4.3\% on GPT2-large. This indicates that for LLM pretraining, the influence of channel-wise outliers is significant, and our quantization method effectively preserves accuracy.

\vskip -0.2in
\subsection{Ablation Study}\label{Subsec: Ablation}
\paragraph{Quantization Error}\label{Subsec: Quantization Error}
We study the quantization error of different quantization methods on four different sizes of GPT2 models to show our method's effectiveness. We focus on the activation of the final layer and calculate the mean squared error (MSE) and the mean error after quantization. The results are shown in Fig.~\ref{Fig: Quantization Error of different methods}. 
For all models, per-channel quantization consistently resulted in smaller quantization errors compared to per-token quantization. Jetfire (ours) achieves lower quantization error than per-token quantization while performing on par with per-channel quantization.
\vskip -0.3in
\paragraph{CUDA kernel and Triton kernel block size}
We have found that the selection of the block size for Triton and CUDA kernels is crucial. A large block size leads to a decrease in parallelism, while a small block size results in low utilization of bandwidth and computational resources. Both cases can result in low kernel speed. In Fig.~\ref{Fig: GELU Block size Ablation}~\ref{Fig: GEMM CUDA}, we test the kernel's speed under different block sizes and find that optimal efficiency is achieved when we set Triton block size$~=64 \times 64$ and CUDA block size$~=128\times 32 \times 128$.

\vskip -0.2in
\subsection{Operator and End-to-End experiments}\label{Subsec: Operator speed}
\paragraph{Linear layer speedup}
We test the speedup of our custom linear layer on RTX 4090. We analyzed the time consumption of each component in forward and backward passes and compared the speed of our implementation with FP16 and SwitchBack linear layers. The results are shown in Fig.~\ref{Fig: Matmul Speed Test}. Our MM \textbf{operator} provides about 60\% speed improvement compared to FP16. Other overhead components like quantizing and reshaping have a relatively minor impact. Our method achieves 40\% overall speedup (forward + backward), which is comparable to the acceleration result of SwitchBack, where SwitchBack leaves the calculation of weight gradient in FP. The speedup becomes larger when the matrix size increases since the overhead proportion decreases, which is demonstrated in Table~\ref{Table: Overhead portion in Linear Layer}. Acceleration results on RTX 3090 can be found in Appendix~\ref{sec: other hardware}.
\vskip -0.2in
\paragraph{Non-linear operator speedup}
We also test the speedups offered by our custom non-linear layers, which is the \textbf{first} quantized training work to achieve acceleration for these non-linear operators.

Our INT8 GELU operator achieves 80\% speedup in both forward and backward passes compared to PyTorch's FP16 operators. Our INT8 LayerNorm operator achieves 40\% speed up in its forward pass and up to 90\% speedup in its backward pass when $\text{hidden size}=8192$ but does not accelerate when the hidden size is small. These results indicate that the global memory access is indeed the bottleneck for these non-linear operators, and our INT8 data flow can effectively solve the bottleneck, resulting in near-ideal speedup.
\vskip -0.2in
\paragraph{End-to-end speedup}
We experimented with GPT2 models and varied the network hidden size to show the end-to-end speedup for our Jetfire method over PyTorch's FP16 training on RTX 4090. We integrated all linear and non-linear operators and reported the speedup of a transformer layer. We compared the forward, backward, and overall runtime speedup with the SwitchBack layer. Results in Table~\ref{Table: End-to-End accelerate} showed that our method achieved comparable or improved acceleration compared to SwitchBack. This is primarily because our linear operators in backpropagation are faster than SwitchBack, and we can accelerate all of the non-linear operators in both forward and backward propagation. Acceleration results on RTX 3090 can be found in Appendix~\ref{Table: Accelerate on 3090}.

\begin{table}[]
\centering
\caption{Acceleration ratios for End-to-end comparison (SB refers to
SwitchBack basic version) on GPT2 model.}
\label{Table: End-to-End accelerate}
\begin{footnotesize}
\begin{sc}
\resizebox{0.9\linewidth}{!}{
\begin{tabular}{ccccccc}
\toprule
~ & 
\multicolumn{2}{c}{Forward} & \multicolumn{2}{c}{Backward} & \multicolumn{2}{c}{Overall} \\ 
\cmidrule(r){2-3} \cmidrule(r){4-5} \cmidrule(r){6-7}
Hidden Size &  SB      & Ours   & SB    & Ours  & SB  & Ours     \\ 
\midrule
4096  & 1.50 & 1.32 & 1.18 & 1.46 & 1.27 & \textbf{1.42} \\
\midrule
2048                         & 1.53    & 1.29   & 1.24  & 1.41  & 1.32 & \textbf{1.37} \\
\midrule
1024                         & 0.94    & 0.97   & 1.14  & 1.11  & \textbf{1.07} & \textbf{1.07}  \\
\bottomrule
\end{tabular}
}
\end{sc}
\end{footnotesize}
\end{table}
\paragraph{End-to-End Memory Reduction}\label{Subsec: Operator Memory}
We experimented with GPT2 models and varied the network depth and batch size to show the memory reduction of our method. We report the reduction ratio of activation memory. The results are shown in Table~\ref{Table: Memory Reduction}. Our method achieved up to 1.49x activation memory reduction, which is better than SwitchBack since we reduced the memory footprint of non-linear operators.

\begin{table}[]
\centering
\caption{Activation memory reduction ratios for End-to-end comparison (SB refers to
SwitchBack Memeory Efficient version) on GPT2 model.}
\label{Table: Memory Reduction}
\begin{footnotesize}
\begin{sc}
\resizebox{0.9\linewidth}{!}{
\begin{tabular}{ccccccc}
\toprule
~ & 
\multicolumn{2}{c}{BS=1} & \multicolumn{2}{c}{BS=2} & \multicolumn{2}{c}{BS=4} \\ 
\cmidrule(r){2-3} \cmidrule(r){4-5} \cmidrule(r){6-7}
Layer Num &  SB      & Ours   & SB    & Ours  & SB  & Ours     \\ 
\midrule
12  & 1.19 & \textbf{1.33} & 1.14 & \textbf{1.31} & 1.11 & \textbf{1.29}
 \\
\midrule
24  & 1.24 & \textbf{1.49} & 1.18 & \textbf{1.47} & 1.14 & \textbf{1.45} \\
\bottomrule
\end{tabular}
}
\end{sc}
\end{footnotesize}
\end{table}
\vskip -0.1in
\section{Conclusion}
\vskip -0.1in
In this work, we propose Jetfire, an INT8 pretraining method for transformer models. For the first time, we propose to use INT8 data flow in pretraining to reduce computation, memory access, memory usage, and communication at the same time. We also propose to use per-block quantization for all of the activations, weights, and gradients for both linear and non-linear layers to preserve accuracy. Extensive experiments demonstrate that our proposed method performs on par with FP baselines, and can effectively accelerate the training speed and reduce the memory footprint. 

\section*{Acknowledgements}
The authors would like to thank Bingrui Li, Ziteng Wang, Jiayi Zhong, Cheng Lu for the helpful discussions. This work was supported by the National Science and Technology Major Project (2021ZD0110502),  NSFC Projects (Nos.~62376131, 62061136001, 62106123, 62076147, U19A2081, 61972224), Tsinghua Institute for Guo Qiang, and the High Performance Computing Center, Tsinghua University. J.Z is also supported by the XPlorer Prize.
\section*{Impact Statement}
Our INT8 fully quantized training (FQT) method significantly improves the efficiency of deep learning by reducing computations and memory usage of training transformers. This contributes substantially to energy conservation and emission reduction, and aligns with the objective of global sustainability. Besides, our method promotes the democratization of artificial intelligence (AI) by making transformer training more accessible to cheap and low-resource platforms. Nevertheless, this method could also be misused to expedite the training of "evil models" designed to generate harmful content.

\newpage
\bibliography{example_paper}
\bibliographystyle{icml2024}

\newpage
\appendix
\onecolumn

\section{Triton Implementation of Non-Linear Operators}
\label{App: Non-Linear}

For the GELU function, its forward and backward operator is: 
\[\mathrm{GELU}(x) = x\cdot \Phi(x),~~
\frac{\mathrm{d} \mathrm{GELU}(x)}{\mathrm{d} x} = \frac{x}{\sqrt{2\pi}}e^{-\frac{x^2}{2}} + \Phi(x).\]
For Dropout, its forward and backward operator is: 
\[
\mathrm{Drop}(x) = \frac{1}{1-p} x \circ m, ~~
\frac{\mathrm{d}\mathrm{Drop}(x)}{\mathrm{d}x} = \frac{1}{1-p}m.
\]
For Add, when we calculate the residual connection $y = x + f(x)$, we also need to perform $\mathrm{d}x = \mathrm{d}f(y) + \mathrm{d}y$ in the backward process. This addition operator can be represented as:
\[
\mathrm{Add}(x_1, x_2) = x_1 + x_2.
\]

\begin{algorithm}
\small
\caption{INT8 Non-Linear Operator}
\label{Alg: Non-linear}
\begin{algorithmic}[1]
\REQUIRE INT8 Matrix $\Xv \in \Rb^{N \times C}$, FP16 scale matrix $\Sv_{\Xv} \in \Rb^{L_N \times L_C}$, element-wise function $f$

\STATE Define $T_N = \left\lceil \frac{N}{B_N} \right\rceil$, $T_C = \left\lceil \frac{C}{B_C} \right\rceil$

\STATE Define $R_N = \left\lceil \frac{B_N}{B} \right\rceil, R_C = \left\lceil \frac{B_C}{B} \right\rceil$

\FOR{1 $\le$ i $\le T_N$}
    \FOR{1 $\le$ j $\le T_C$}
    \STATE Load INT8 block $\Xv_{ij}\in\Rb^{B_N \times B_C}$, $\Sv_{\Xv_{ij}}\in\Rb^{R_N \times R_C}$
    \STATE Dequantize $\Xv_{ij}$ and $\Sv_{\Xv_{ij}}$ to get $\Xv_{ij}^{\textbf{FP32}}$
    \STATE Operate: $\Yv_{ij}^{\textbf{FP32}} = f(\Xv_{ij}^{\textbf{FP32}})$
    \STATE Quantize $\Yv_{ij}^{\textbf{FP32}}$ to get $\Yv_{ij}^{\textbf{INT32}} \in \Rb^{B_N \times B_C}$ and scale factor $\Sv_{\Yv_{ij}} \in \Rb^{R_N \times R_C}$
    \STATE Save $\Yv_{ij}^{\textbf{INT32}}$ and $\Sv_{\Yv_{ij}}$ to global memory.
    \ENDFOR
\ENDFOR
\end{algorithmic}
\end{algorithm}

Differing from non-linear operators above, LayerNorm involves interactions between elements. Therefore, performing calculations separately for each $B_N \times B_C$ block is not feasible. In order to solve the problem, we observed that both pre-norm and post-norm models encountered the ADD operator before LayerNorm. 

We make the following modifications to our ADD operator: We will calculate the mean and sum of squares for each row of the block $(B_N, B_C)$ and store these values. We will then get the mean matrix and sum of squares matrix of size $N \times \frac{C}{B_C}$, which reduces the amount of data we need to load and store by $\frac{1}{B_C}$. Before the LayerNorm operator, we use these values to compute the mean and variance for each row, which size is $N\times 1$. This allows the LayerNorm to directly access these values. The implementation of the remaining part of LayerNorm is similar to that of GELU.
\newpage
\section{Detailed Results of GLUE Fine-Tuning Test}
\label{App: Detailed GLUE Results}

\begin{table*}[h]
\caption{Detailed Results of GLUE fine-tuning test based on the pretrained model. FP refers to floating-point, SwitchBack refers to per-token quantization. '--' means the model does not converge.}
\begin{center}
\begin{footnotesize}
\begin{sc}
\resizebox{0.9\linewidth}{!}{
\begin{tabular}{ccccccc}
\toprule
~ & ~ & ~ & \multicolumn{3}{c}{Baseline} &
Ours
\\
\cmidrule(r){4-6} \cmidrule(r){7-7}
Model & $\#$Params(M) & METRIC & FP & Switchback & Per-Tensor & Jetfire
\\
\midrule
\multirow{8}{*}{GPT2-BASE} & \multirow{8}{*}{124} &
     COLA & $43.97_{3.32}$ & $\boldsymbol{44.68_{1.06}}$ & $41.43_{2.23}$ & $41.98_{2.51}$ \\
& &  STSB & $83.20_{0.95}$ & $81.08_{1.39}$ & $75.63_{5.35}$ & $\boldsymbol{81.66_{0.45}}$ \\
& &  RTE  & $64.02_{1.37}$ & $64.74_{0.83}$ & $64.14_{0.75}$ & $\boldsymbol{65.22_{0.91}}$ \\
& &  MRPC & $85.48_{0.78}$ & $84.30_{0.49}$ & $83.68_{0.61}$ & $\boldsymbol{85.01_{0.49}}$ \\
& &  SST2 & $91.40_{0.30}$ & $\boldsymbol{92.09_{0.50}}$ & $90.79_{0.26}$ & $91.93_{0.35}$ \\
& &  QNLI & $87.97_{0.35}$ & $\boldsymbol{87.58_{0.19}}$ & $86.64_{0.25}$ & $87.57_{0.19}$ \\
& &  QQP  & $90.07_{0.001}$& $89.94_{0.07}$ & $88.84_{0.09}$ & $\boldsymbol{90.14_{0.03}}$ \\
& &  MNLI & $81.84_{0.22}$ & $80.76_{0.24}$ & $79.51_{0.36}$ & $\boldsymbol{81.92_{0.18}}$ \\
\midrule
\multirow{8}{*}{GPT2-MEDIUM} & \multirow{8}{*}{350} &
     COLA & $53.55_{2.89}$ & $50.97_{0.58}$ & $48.82_{1.43}$ & $\boldsymbol{53.38_{1.08}}$ \\
& &  STSB & $86.17_{0.78}$ & $83.21_{1.61}$ & $81.66_{0.50}$ & $\boldsymbol{84.98_{0.32}}$ \\
& &  RTE  & $68.23_{1.44}$ & $63.78_{0.91}$ & $67.27_{0.21}$ & $\boldsymbol{68.47_{0.21}}$ \\
& &  MRPC & $88.16_{0.91}$ & $84.79_{0.68}$ & $84.98_{0.15}$ & $\boldsymbol{86.74_{1.36}}$ \\
& &  SST2 & $93.73_{0.29}$ & $\boldsymbol{94.00_{0.18}}$ & $91.78_{0.13}$ & $93.46_{0.23}$ \\
& &  QNLI & $90.49_{0.14}$ & $89.61_{0.20}$ & $88.05_{0.16}$ & $\boldsymbol{90.15_{0.17}}$ \\
& &  QQP  & $90.98_{0.15}$ & $90.54_{0.15}$ & $90.02_{0.08}$ & $\boldsymbol{90.93_{0.08}}$ \\
& &  MNLI & $84.69_{0.37}$ & $83.37_{0.36}$ & $80.93_{0.31}$ & $\boldsymbol{84.65_{0.11}}$ \\
\midrule
\multirow{8}{*}{GPT2-LARGE} & \multirow{8}{*}{774} &
     COLA & $52.22_{0.15}$ & $48.70_{1.20}$ & $35.85_{1.23}$ & $\boldsymbol{54.06_{1.62}}$ \\
& &  STSB & $84.99_{1.37}$ & $79.49_{0.15}$ & $80.00_{0.15}$ & $\boldsymbol{84.99_{1.37}}$ \\
& &  RTE  & $76.65_{1.67}$ & $65.10_{1.10}$ & $64.26_{2.60}$ & $\boldsymbol{74.73_{2.53}}$ \\
& &  MRPC & $86.70_{0.92}$ & $83.75_{1.00}$ & $83.09_{0.45}$ & $\boldsymbol{87.06_{1.21}}$ \\
& &  SST2 & $94.84_{0.50}$ & $92.47_{0.07}$ & $90.86_{0.24}$ & $\boldsymbol{94.65_{0.18}}$ \\
& &  QNLI & $91.35_{0.23}$ & $88.42_{0.12}$ & $85.25_{0.36}$ & $\boldsymbol{91.19_{0.52}}$ \\
& &  QQP  & $91.40_{0.08}$ & $89.92_{0.19}$ & $89.21_{0.03}$ & $\boldsymbol{91.19_{0.02}}$ \\
& &  MNLI & $85.88_{0.14}$ & $82.05_{0.20}$ & $78.53_{0.15}$ & $\boldsymbol{85.61_{0.30}}$ \\

\bottomrule
\end{tabular}}
\end{sc}
\end{footnotesize}
\end{center}
\vskip -0.1in
\end{table*}
\section{Comparisons with methods targeting CNNs}\label{Sec: More baselines}
In this section, We tested two INT8 training for CNN models~\cite{wang2023GDA, zhao2021DAQ} on the DeiT pretraining experiment. As reported in Table~\ref{Table: More Baselines}, both of them showed significant accuracy degradation. This indicates that these methods are not sufficient to be applied to transformer models.

\begin{table}[h!]
    \centering
    \begin{tabular}{lcccccc}
        \toprule
        Model & FP & SwitchBack & Per-tensor & Jetfire & GDA & DAQ \\
        \midrule
        Deit-tiny  & 64.08 & 63.55 & -    & 63.95 & 62.14 & 61.80 \\
        Deit-small & 73.43 & 72.80 & -    & 73.31 & 70.98 & 70.66 \\
        Deit-base  & 75.67 & 75.62 & -    & 76.03 & 73.06 & 72.40 \\
        \bottomrule
    \end{tabular}
    \caption{Comparison of different methods on various Deit models}
    \label{Table: More Baselines}
\end{table}

\section{Acceleration Experiments}
\subsection{Overhead portion in Linear Layer} 
We tested the percentage of time taken by all quantization, dequantization, transpose, and other overhead processes during the forward and backward passes in a linear layer. We find that in Table~\ref{Table: Overhead portion in Linear Layer}, the relative overhead from quantization/dequantization diminishes with increasing model size, leading to more significant speed improvements.

\begin{table}[t]
\centering
\begin{sc}
\begin{tabular}{ccccccc}
\toprule
~ & ~ & \multicolumn{5}{c}{Sequence Length} \\
\cmidrule(r){3-7}
Batch Size & Place & 1024 & 2048 & 4096 & 6144 & 8192 \\
\midrule
4 & forward & 44.5 & 22.1 & 10.2 & 6.8 & 5.5 \\
4 & backward & 30.4 & 29.0 & 16.6 & 12.2 & 9.2 \\
4 & overall & 34.5 & 26.7 & 14.5 & 10.3 & 7.9 \\
16 & forward & 23.1 & 8.1 & 2.7 & 1.7 & 1.3 \\
16 & backward & 24.6 & 26.4 & 8.7 & 7.3 & 6.2 \\
16 & overall & 23.7 & 18.3 & 11.5 & 9.9 & 8.9 \\
\bottomrule
\end{tabular}
\end{sc}
\caption{Percentage of overhead in a linear layer.}
\label{Table: Overhead portion in Linear Layer}
\end{table}

\subsection{Acceleration result on other hardware}\label{sec: other hardware}
Besides RTX 4090, we tested our linear operator and end-to-end speed up result on the RTX 3090 GPUs, as reported in Table~\ref{Table: Accelerate on 3090}. The results indicate that our method can achieves significant speedups on multiple kinds of GPUs.

\begin{table}[t]
\centering
\begin{sc}
\begin{tabular}{ccccccc}
\toprule
\textbf{Size Settings} & \textbf{Fwd-SB} & \textbf{Fwd-Ours} & \textbf{Bwd-SB} & \textbf{Bwd-Ours} & \textbf{All-SB} & \textbf{All-Ours} \\
\midrule
Linear Layer, C=D=2048 & 1.60 & 1.53 & 1.38 & 1.31 & 1.45 & 1.38 \\
Linear Layer, C=D=4096 & 2.43 & 1.87 & 1.37 & 1.48 & 1.62 & 1.60 \\
Linear Layer, C=D=8192 & 2.56 & 1.70 & 1.24 & 1.40 & 1.51 & 1.49 \\
\midrule
End-to-End, hidden=1024 & 1.08 & 0.94 & 1.08 & 1.10 & 1.08 & 1.05 \\
End-to-End, hidden=2048 & 1.34 & 1.18 & 1.15 & 1.36 & 1.21 & 1.29 \\
End-to-End, hidden=4096 & 1.27 & 1.23 & 1.18 & 1.37 & 1.24 & 1.32 \\
\bottomrule
\end{tabular}
\end{sc}
\caption{Speed up result on the RTX 3090 GPUs. SB refers to SwitchBack, Ours refers to Jetfire.}
\label{Table: Accelerate on 3090}
\end{table}

\end{document}